\documentclass{bmvc2k}

\usepackage{hyperref}       
\usepackage{url}            
\usepackage{booktabs}       
\usepackage{amsfonts}       
\usepackage{nicefrac}       
\usepackage{microtype}      

\usepackage{graphicx}
\usepackage{subfigure}

\usepackage{mathtools}
\DeclarePairedDelimiter{\ceil}{\lceil}{\rceil}
\DeclareMathOperator{\Pred}{Pred}
\DeclareMathOperator{\Nbr}{\cal N}
\usepackage{stmaryrd}
\usepackage{relsize}
\usepackage{color}
\usepackage{stackengine}
\usepackage{amsthm}
\usepackage{amssymb}

\newtheorem{thm}{Theorem}
\newtheorem{definition}{Definition}

\def\ie{\emph{i.e.}}

\definecolor{darkgreen}{RGB}{0,150,0}

\graphicspath{ {Figures/} }

\title{The Resistance to Label Noise in $K$-NN and DNN Depends on its Concentration\footnote{Accepted to BMVC 2020}}

\addauthor{Amnon Drory}{amnondrory@mail.tau.ac.il}{0}
\addauthor{Oria Ratzon}{ oriaratzon1@gmail.com}{0}
\addauthor{Shai Avidan}{avidan@eng.tau.ac.il}{0}
\addauthor{Raja Giryes}{raja@tauex.tau.ac.il}{0}

\addinstitution{
 Tel-Aviv University\\
  Israel
}

\runninghead{Drory, Ratzon, Avidan, Giryes}{Label Noise in $K$-NN and DNN}

\def\etal{\emph{et al}\bmvaOneDot}

\begin{document}

\maketitle

\begin{abstract}
We investigate the classification performance of $K$-nearest neighbors ($K$-NN) and deep neural networks (DNNs) in the presence of label noise. We first show empirically that a DNN’s prediction for a given test example depends on the labels of the training examples in its local neighborhood. This motivates us to derive a realizable analytic expression that approximates the multi-class $K$-NN classification error in the presence of label noise, which is of independent importance. We then suggest that the expression for $K$-NN may serve as a first-order approximation for the DNN error. Finally, we demonstrate empirically the proximity of the developed expression to the observed performance of $K$-NN and DNN classifiers. Our result may explain the already observed surprising resistance of DNN to some types of label noise. It also characterizes an important factor of it showing that the more concentrated the noise the greater is the degradation in performance.
\end{abstract}


\section{Introduction} \label{sec:introduction}
Deep neural networks (DNN) provide state-of-the-art results in many computer vision challenges, such as image classification \cite{NIPS2012_4824}, detection \cite{redmon2016yolo9000} and segmentation \cite{DBLP:journals/corr/ChenPK0Y16}. Yet, to train these models, large datasets of labeled examples are required. Time and cost limitations come into play in their creation, which often result in imperfect labeling, or \emph{label noise}, due to human error \cite{Ipeirotis10Quality}. An alternative to manual annotation are images taken from the Internet that use the surrounding text to produce labels \cite{Veit2017Learning}. This approach results in noisy labels too.

\begin{figure}
\centering
  \includegraphics[width=0.25\linewidth]{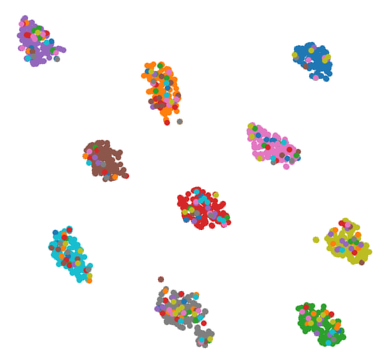} \hspace{0.3in}
\includegraphics[width=0.25\linewidth]{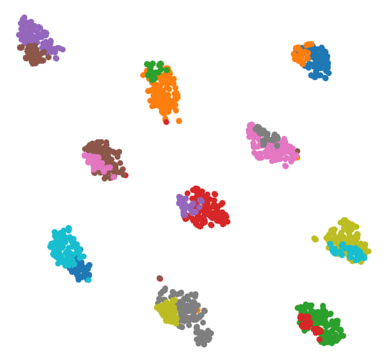}
\caption{\small {\bf Effect of label noise type:} Each cluster represents a class and the color represents the label provided for each data point. (left) $20\%$ uniformly spread random noise. The network achieves $\sim100\%$ prediction accuracy. (right) $20\%$ locally concentrated noise. The network achieves $\sim80\%$ accuracy.}
\label{fig:NoiseTypes}
\end{figure}

Perhaps surprisingly, it has been shown \cite{2015arXiv151106789K} that DNNs trained on datasets with high levels of label-noise may still attain accurate predictions. This phenomenon is not unique only to DNNs but also to classic classifiers such as $K$-Nearest Neighbours ($K$-NN) \cite{Angluin88Learning, Natarajan13Learning, Frenay14Classification}.

Fig.~\ref{fig:NoiseTypes} demonstrates this behavior. It shows the 10 MNIST classes, using deep features embedded in 2-dimensional space with t-SNE~\cite{vanDerMaaten2008}. We change the labels of a randomly selected 20\% of the training data. A network trained with this noisy data is capable of reaching $~100\%$ prediction accuracy. On the other hand, it also shows the case where concentrated groups of examples have all their labels flipped to the same label. Here too, 20\% are changed, but the noise is no longer distributed uniformly in the example set space but is instead is locally concentrated. In this case, the DNN does not overcome the noise and accuracy drops to $~80\%$. Clearly, the \emph{type} of label noise is as important as its \emph{amount}.

The starting point of this work is the observation that a DNN's prediction for any given test example depends on a local \emph{neighborhood} of training examples, where \emph{local} is in some implicit space. We indirectly demonstrate this by injecting varying levels of a known noise into the training set, and measuring the effect - first on the softmax output of the network, and secondly on accuracy-vs-noise-level curves.

The output of a classification DNN's last-layer, the softmax layer, is a probability distribution over the possible labels. We show that when noise is added, these distributions tend to encode the distribution of labels of training examples in a local neighborhood around a test example. This means that the output of a DNN is similar to the output of a $K$-NN algorithm performed in some implicit space. Motivated by this, we are able to develop an analytical expression for the expected accuracy of a $K$-NN algorithm given a noisy training set, when the parameters of the noise producing process are known. This allows us to produce analytical accuracy-per-noise-level curves. Next, we actually produce noisy training sets with varying levels of noise, train a DNN on each, and measure its accuracy. This results in \emph{experimental} accuracy-per-noise-level curves for DNNs. By comparing the analytical and experimental curves, we demonstrate that the $K$-NN model is a good first order prediction for the behaviour of DNNs in the presence of noise. This is an important step forward in understanding DNN's resistance to label noise. It also allows us to recognize a very important factor governing the extent to which DNNs can resist label noise: whether the noise is \emph{randomly spread} or \emph{locally concentrated} in the training set. When label noise is \emph{randomly spread}, the resistance to noise is high, since the probability of noisy examples overcoming the correct ones in any local neighborhood is small. However, when the noisy examples are \emph{locally concentrated}, DNNs are unable to overcome the noise. 

Our findings are also motivated by recent works that show that networks perform a smooth interpolation between the labels of training examples \cite{Ongie2020Function,Savarese19How,Williams19Gradient,giryes2020function}. We suggest using  $K$-NN as a first order approximation of such an interpolation.

We validate our analytical expression both for $K$-NN and DNN using extensive experiments on several datasets: MNIST, CIFAR-10, and ImageNet ILSVRC.\footnote{Code is available at https://github.com/AmnonDrory/CNNs-With-Label-Noise}
We show that empirical curves of accuracy-per-noise-level fit well with our mathematical expression.

\section{Related Work} \label{sec:relatedWork}

Classification in the presence of label noise has long been explored in the context of classical machine learning \cite{Angluin88Learning, Natarajan13Learning, Frenay14Classification}.
We focus on $K$-NN robustness and its implication on DNN.

{\bf $K$-NN label-noise robustness.} $K$-NN sensitivity to label noise has been discussed in multiple works  \cite{Sanchez:1997:PSN:269573.269576,Wilson:2000:RTI:343196.343200}. Prasath et al. \cite{DBLP:journals/corr/abs-1708-04321} show an impressive resistance of $K$-NN to uniform label noise: drop of only 20\% for 90\% noise level.  
Tomašev and Buza \cite{TOMASEV2015157} observe that resistance to noise depends on its type. They present the \emph{hubness-proportional} noise model, where the probability of an example being corrupted depends not only on its label, but also on its nearness to other examples, and show that this noise is harder for $K$-NN than uniform noise.  

The closest $K$-NN theory to ours is  \cite{Okamoto:1995:AAK:646264.685911}, which provides an expression for the $K$-NN accuracy as a function of noise level. Their derivation relies on the particulars of a specific setting, namely a binary classification task where the inputs are binary strings and the output is 1 if a majority of relevant places in the string are set to 1. Our work can be seen as expanding and generalizing their derivation, as it handles: (i) general input domains; (ii)  multi-class classification (not just 2 classes); and (iii) a much more complex family of noise models.

{\bf DNN label-noise robustness.} The effect of label noise on neural networks has been studied as well. Several works, e.g. \cite{Flatow17robustness,2015arXiv151106789K,SunSSG17, DBLP:journals/corr/RolnickVBS17} have shown that neural networks trained on large and noisy datasets can still produce highly accurate results. For example, Krause {\em et al.} \cite{2015arXiv151106789K} report classification results on up to $10,000$ categories. Their key observation is that working with large scale datasets that are collected by image search on the web leads to excellent results even though such data is known to contain noisy labels. 




Sun {\em et al.} \cite{SunSSG17} report logarithmic growth in performance as a function of training set size. They perform their experiments on the JFT-300M dataset, which has more than 375M noisy labels for 300M images. The annotations have been cleaned using complex algorithms. Still, they estimate that as much as $20\%$ of the labels are noisy and they have no way of detecting them. Wang {\em et al.} \cite{Wang18FaceNoise} investigate label-noise in face recognition and its impact on accuracy.

In \cite{DBLP:conf/icassp/BekkerG16,Goldberger17Training}, an extra noise layer is introduced to the network to address label noise. It is assumed that the observed labels were created from the true labels by passing through a noisy channel whose parameters are unknown. They simultaneously learn the  network parameters and the noise distribution. 
Another approach \cite{DBLP:conf/cvpr/XiaoXYHW15, Vahdat17Toward, Li17Learning, Khetan18LEARNING} 
models the relationships between images, class labels and label noise using a probabilistic graphical model and further integrate it into an end-to-end deep learning system. 
Other methods that show that training on additional noisy data may improve the results appear in 
\cite{Lee18CleanNet, Guo18CurriculumNet, Ding18Semi, Litany18SOSELETO, Li2018LearningTL}.  

Several methods ``clean-up'' the labels by analyzing given noisy labeled data \cite{Tanaka18Joint, Ren18Learning, han2018coteaching, han2018masking, Jiang18MentorNet, Thulasidasan2019CombatingLN, Shen19Learning, Konstantinov2019RobustLF}.
For example, Reed {\em et al.} \cite{DBLP:journals/corr/ReedLASER14} combat noisy labels by means of consistency. They consider a prediction to be consistent if the same prediction is made given similar percepts, where the notion of similarity is between deep network features computed from the input data. 
Malach and Shalev-Schwartz \cite{Malach17Decoupling} suggest a different method for overcoming label noise. They train two networks, and only allow a training example to participate in the stochastic  gradient descent stage of training if these networks \emph{disagree} on the prediction for this example. This allows the training process to ignore incorrectly labeled training examples, as long as both networks agree about what the correct label should be.

Liu \etal \cite{Liu16Classification} propose to use importance reweighting to deal with label noise in DNN. They  extend the idea of using an unbiased loss function for reweighting to improve resistance to label noise in the classical machine learning setting \cite{Angluin88Learning, Natarajan13Learning,Frenay14Classification}. Another strategy suggests to employ robust loss functions to improve the resistance to label noise \cite{Ghosh15Making,Ghosh17Robust, Zhang18Generalized, Jindal16Learning, Patrini17Making, Kaneko19Label}.
In \cite{Yao19Deep} an extra variable is added to the network to represent the trustworthiness of the labels, which helps improving the training with noisy labels.  
Ma {\em et al.} \cite{Ma18Dimensionality} study the dimensionality of the learned representations of examples with clean and noisy labels. They show that there is a difference in the dimensionality between the two cases and use it to improve the training. This work studies the embedding space of trained networks and show a relationship to $K$-NN. We use this relationship to provide a first-order estimate for DNN resistance to label noise.






\section{Analysis of Robustness to Label Noise} \label{sec:analysis}

We take the following strategy to analyze label noise: First, we establish the different label noise models to consider. Then we show empirically that the output of the DNN's softmax resembles the label distribution of the $K$ nearest neighbors, linking DNNs to the $K$-NN algorithm. With this observation in hand, we derive a formula for $K$-NN, which is of interest by itself, with the hypothesis that it applies also to DNN. 

\noindent {\bf Setting.} In the ``ideal'' classification setting, we have a training set $\mathcal{T}=\{x_i, y_i\}_{i=1}^N$ and a test set $\mathcal{S}=\{\hat{x}_i, \hat{y}_i\}_{i=1}^M$, where $x$ is typically an image, and $y$ is a label from the label set $\mathcal{L}=\{\ell_1, \ell_2, \ldots, \ell_L\}$. A classification algorithm (DNN or $K$-NN) learns from $\mathcal{T}$ and is tested on $\mathcal{S}$.  The setting with label noise is similar, except that the classifier learns from a \emph{noisy} training set $\{x_i, \tilde{y}_i\}_{i=1}^N$, which is derived from the clean data $\mathcal{T}$ by changing some of the labels. We designate by $\gamma$ the fraction of training examples that are \emph{corrupted} (with changed labels).\footnote{Note that the subset of "corrupted" examples may actually contain examples whose label has not changed. This happens when the randomly selected noisy label happens to be the same as the original label.} 

An often studied noise setting is that of \emph{randomly-spread} noise. In this setting the process of selecting the noisy label $\tilde{y}$ is agnostic to the content of the image $x$, and instead only depends (stochastically) on the clean label $y$. The examples that get corrupted (i.e. their labels are changed) are selected uniformly at random from the training set $\mathcal{T}$. For each such example, the noisy label is stochastically selected according to a conditional probability $P(\tilde{y}|y)$ (we refer to this as the \emph{corruption matrix}).\footnote{A \emph{confusion matrix} $C$ can be derived from the corruption matrix by  $C=(1\mkern-4mu-\mkern-2mu\gamma)I + \gamma P$, where $I$ is the identity matrix.} This setting can capture the overall similarity in appearance between categories of images, which leads to error in labeling. 

Two simple variants of randomly spread noise are often used: Uniform Noise, and Flip-Noise.
\emph{Uniform Noise} is the case where the noisy label is selected uniformly at random from $\mathcal{L}$. This corresponds to a corruption matrix where $P(\tilde{y}|y)=\frac{1}{L}$ for all $\tilde{y},y$. In the \emph{flip label-noise} setting, each label $\ell_i$ has one counterpart $\ell_j$ with which it may be replaced. In this case each row in the corruption matrix has exactly one entry that is 1, and the rest are 0.

In contrast with the \emph{randomly spread} setting, we also consider the \emph{locally concentrated noise} setting, where the noisy labels are locally concentrated in the training set \cite{Inouye17Hyperparameter}. As an example, consider a task of labeling images as either \emph{cat} or \emph{dog}, and a human annotator that consistently marks all poodles as \emph{cat}. 
We show that $K$-NN and, by extension DNN, are resilient to randomly spread label noise but not to locally concentrated one.

\subsection{The Connection Between DNN and $K$-NN.}
We observe that DNN's prediction, similar to $K$-NN, tends to be the \emph{plurality label} (the most common label) in a local neighborhood of train examples that surround the test example. The connection between $K$-NN and DNN is observed indirectly, by adding different types of noise to the training set, and analyzing its effect on the network's \emph{softmax-layer output}. We find that this output tends to be the local probability distribution of the training examples in the vicinity of $x$. Therefore, its argmax $\ell_{pred}=\displaystyle{\arg \max_{\ell \in \mathcal{L}} \{softmax_x(\ell)\}}$ tends to be the plurality label.

\begin{figure}
\centering
\begin{subfigure}[CIFAR-10, 30\% uniform noise]{
  \includegraphics[width=0.21\linewidth]{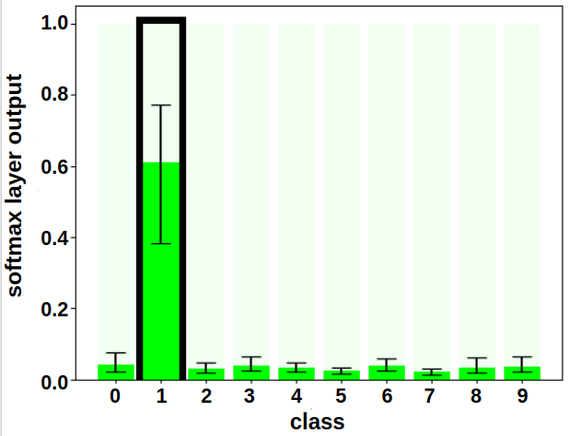}}
  \label{fig:bar_uniform_60}
\end{subfigure}
\hfill
\begin{subfigure}[CIFAR-10, 40\% flip noise]{\includegraphics[width=0.21\linewidth]{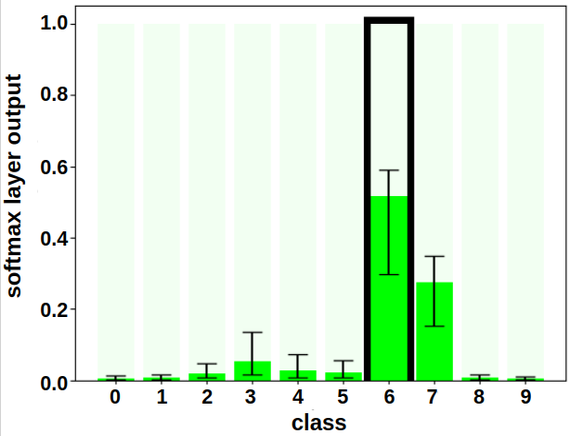}}
\label{fig:bar_flip_20}
\end{subfigure}
\hfill
\begin{subfigure}[MNIST, Locally concentrated noise, example in a clean region]{
  \includegraphics[width=0.21\linewidth]{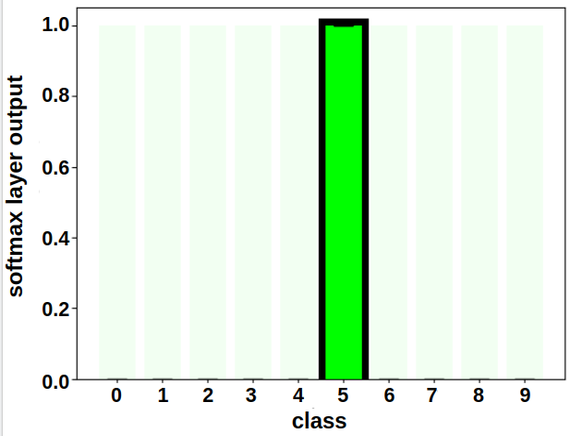}}
\label{fig:mnist_locally_concentrated_025_clean_region}
\end{subfigure}
\hfill
\begin{subfigure}[MNIST, Locally concentrated noise, example in a noisy region]{\includegraphics[width=0.21\linewidth]{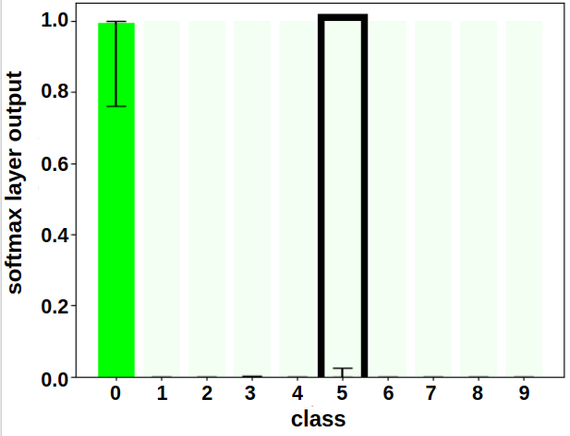}}
\label{fig:mnist_locally_concentrated_025_noisy_region}
\end{subfigure}
\caption{\small {\bf Softmax analysis:} 
Each diagram is aggregated from many test examples. The height of the bars shows the median, and the confidence interval shows the central 50\% of examples. The ground truth label is marked by a black margin. 
Additional softmax diagrams are found in the appendix.}
\label{fig:softmax}
\end{figure}

Fig.~\ref{fig:softmax} presents the average softmax output of DNNs for various noise types and datasets. It demonstrates how the softmax layer output tends to be the distribution of the labels in the neighborhood of training examples. For example, when there is a uniform noise with noise level $\gamma$, we see that the value of the softmax is approximately $\frac{\gamma}{L}$ everywhere except at the peak location. This is indeed the expected fraction of noisy examples from each class in any given local neighborhood. In the case of flip noise, it can be seen that the softmax probabilities are mostly concentrated at the correct class and the alternative class, and that the alternative class probability is approximately at the noise level. Additional softmax diagrams for MNIST and CIFAR-10 can be found in the appendix. 

It follows that the network makes a wrong prediction only when the ``wrong'' class achieves plurality in a local neighborhood. This, for example, is the case when locally concentrated noise is added and the test example is taken from the noisy region. 

These findings provide us with an intuition into how DNNs are able to overcome label noise: Only the \emph{plurality label} in a neighborhood determines the output of the network. Therefore, adding label noise in a way that does not change the plurality label should not affect the network's prediction. As long as the noise is \emph{randomly spread} in the training set, the plurality label is likely to remain unchanged. The higher the noise level, the more likely it is that a \emph{plurality label switch} will occur in some neighborhoods. 
When the noise type and noise level are known, we are able to produce a mathematical expression for the $K$-NN model, that approximately predicts the probability of a switch. We suggest that this can serve as a first-order approximation to the behaviour of DNNs in the presence of noise. We show empirically that indeed it does match the observed behaviour of DNNs quite well in some settings. We also believe that this expression for $K$-NN is of independent interest, as it improves and extends previously known mathematical models for the resistance of $K$-NNs to noise \cite{Okamoto:1995:AAK:646264.685911}.

For the most part, our analysis does not require us to explicitly specify the space in which distance between examples is defined. Our mathematical derivation of expected accuracy-per-label-noise curves does not rely on any of these specifics, but instead only on the parameter $K$, and knowledge of the noise producing process. The fact that the analytical curves match quite well with the experimental curves for DNN, demonstrates the relation between $K$-NN and DNN. Similarly, no explicit specification of the space is needed to demonstrate the effect of the noise on the DNN's softmax layer output. 

A specific embedding space of interest is the output of the penultimate layer of a trained network, since distances between examples in this space are expected to carry some semantic meaning. In the appendix we describe an experiment that demonstrates the similarity between $K$-NN in this space, and softmax layer outputs.

\subsection{$K$-NN Accuracy in The Randomly-Spread Noise Setting.} 

We turn to produce an analytical expression for the  probability of a plurality switch, in the \emph{randomly-spread} noise setting. 

We model randomly spread noise as follows: each test example $(\hat{x}_s, \hat{y}_s)$ has a local neighborhood $\Nbr(\hat{x}_s)$ of $K$ training examples. $q_i$ is the probability for any example in $\Nbr(\hat{x}_s)$ to have the \emph{observed} label $\ell_i$. The distribution $q$ encodes the results of the noise-creation process, and it depends on the parameters of this process, and on the clean labels of the examples in $\Nbr(\hat{x}_s)$. Following \cite{Okamoto:1995:AAK:646264.685911} we considerably simplify our derivation by introducing a small approximation: instead of treating the clean training examples as constant, we consider them to be sampled i.i.d from a \emph{clean distribution} $C_s(\ell)$.  
The $K$-NN algorithm's prediction, which we denote by $Y(\hat{x}_s)$, is the plurality label. Its expected accuracy is defined as follows.
\begin{definition}[$K$-NN Prediction Accuracy]
$K$-NN prediction accuracy is defined as
\begin{equation}
A_{K-NN} \triangleq \frac{1}{M} \sum_{s=1}^{M} \Pr \big( Y(\hat{x}_s) = \hat{y}_s \big),
\label{eq:A_K-NN}
\end{equation}
where $\Pr \big( Y(\hat{x}_s) = \hat{y}_s \big)$ is the probability that the plurality label of test example $\hat{x}$ in $\Nbr(\hat{x})$ is the same as the ground truth label for $\hat{x}$.
\end{definition}

By expanding the expression in Eq. \eqref{eq:A_K-NN}, we obtain an analytical formula for the accuracy of a $K$-NN classifier, which is given in the following theorem (proof in the sup. material):

\begin{thm}[Plurality Accuracy]
The probability of the plurality label being correct is
\small
\begin{eqnarray}
\label{eq:Q}
 && \hspace{-0.3in} Q \triangleq  \Pr \big( Y(\hat{x}) = \hat{y} \big)  
 = \sum_{n_1}\sum_{n_2}\cdots  \sum_{n_L} \,\llbracket n_i \mkern-4mu > \mkern-4mu n_j,\, \forall j \mkern-4mu \neq \mkern-4mu i \rrbracket  \cdot \binom{K}{n_1,n_2,\ldots,n_L}\cdot q_1^{n_1}\cdots q_L^{n_L},
\end{eqnarray}
\normalsize
where $\llbracket\cdot\rrbracket$ is the indicator function, ${\hat y} = \ell_i$ is the correct label, $n_j$ is the number of appearances of the label $\ell_j$ in $\Nbr(\hat{x})$ and $q_j$ is the probability of any such appearance. 
\end{thm}

What is left to show is how to calculate $q_j$. The probability $q_j$ is derived from the process that creates the noisy training set. Let $\hat{x}_s$ be a test example, and let $x$ be a training example in $\Nbr(\hat{x}_s)$. Let $y$ be the clean label of $x$ and $\tilde{y}$ be its noisy label. We denote by $C_s(\ell)$ the \emph{clean label distribution} in $\Nbr(\hat{x}_s)$. In other words, $C_s(\ell) \triangleq Pr(y=\ell)$. Thus, the expression for $q_j$ is
\begin{eqnarray}
q_j  \triangleq   \Pr (\tilde{y}=\ell_j)  =  (1\mkern-4mu-\mkern-2mu\gamma)\cdot C_s(\ell_j) + \gamma \cdot \sum_{k=1}^{L}P(\ell_j|\ell_k)\cdot C_s(\ell_k),
\label{eq:q_j}
\end{eqnarray}
where $\gamma$ is the noise level, and $P(\tilde{y}|y)$ is the corruption matrix that defines the corruption process. Eq.~\eqref{eq:q_j} shows that an example may be labeled with a noisy label $\ell$ in two ways: Either this example is uncorrupted and $\ell$ was its original label, or this example was corrupted and received $\ell$ as its noisy label. 

We can greatly improve the efficiency of calculating $Q$ by first decomposing the multinomial coefficient into a product of binomials, and then decomposing $Q$ into
\begin{align}
 Q &= \sum_{n_1=m_1}^{M_1}{\binom{K}{n_1}}q_1^{n_1}
\cdots \mkern-9mu \sum_{n_L=m_L}^{M_L}{\binom{K-\sum\limits_{j=1}^{L-1}n_j}{n_L}}q_L^{n_L},
\label{eq:efficient_nested_summation}
\end{align}
where $m_i$ is the smallest number of repeats of $\ell_i$ allowed, $M_i$ is the largest, and together they encode the requirement that $n_i > n_j \,\, \forall j \neq i$.
See appendix for a detailed derivation. Equation~\eqref{eq:efficient_nested_summation} contains many partial sums that are repeated multiple times, which allows further speedups by dynamic programming. 

\begin{figure}
\centering
\begin{subfigure}[MNIST flip]{
\includegraphics[width=0.23\textwidth]{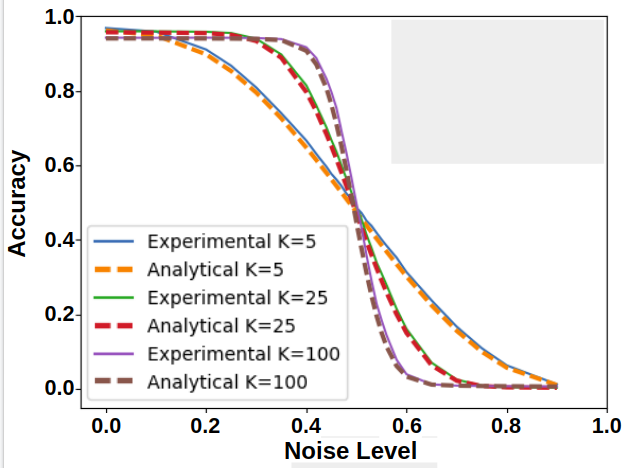}}
\end{subfigure}
\begin{subfigure}[CIFAR-10 flip]{
\includegraphics[width=0.23\textwidth]{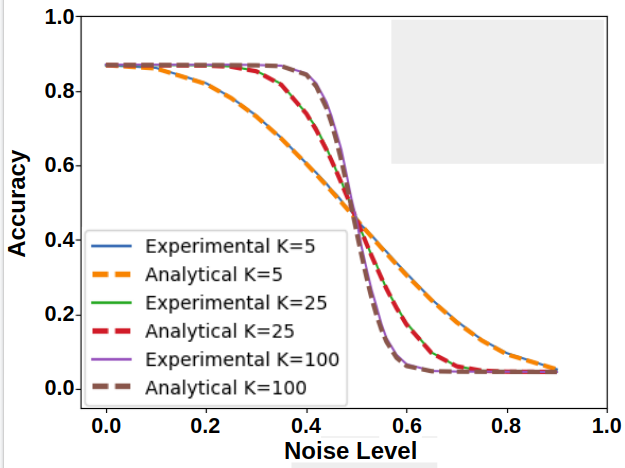}}
\end{subfigure}
\begin{subfigure}[CIFAR-10 uniform]{
\includegraphics[width=0.23\textwidth]{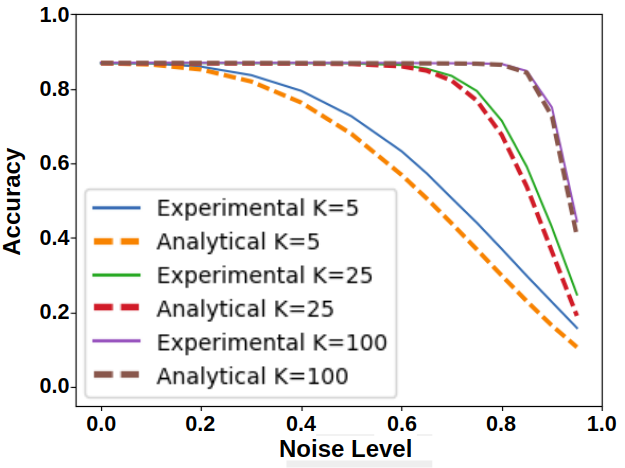}}
\end{subfigure}
\begin{subfigure}[ImageNet uniform]{
\includegraphics[width=0.23\textwidth]{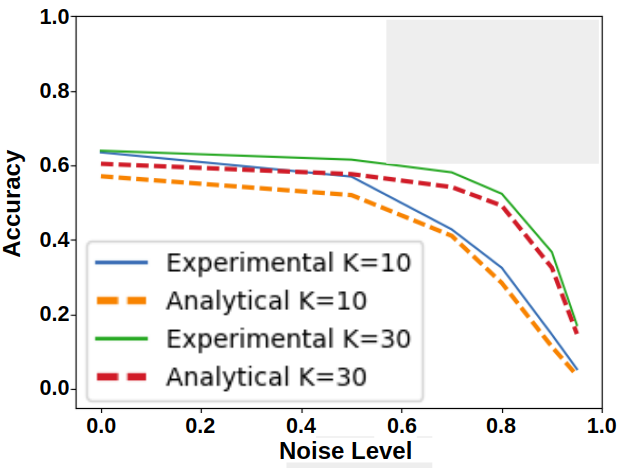}}
\label{fig:KNN_ImageNet_Uniform}
\end{subfigure}
\caption[]
{\small $K$-Nearest Neighbors Analytical and Experimental curves, showing that the effect of noise on accuracy is  predicted very well by the analytical model. For the MNIST dataset, K-NN is performed in the space of image pixels (784 dimensions). For the CIFAR-10 dataset, K-NN is performed in a 256-dimensional feature space, derived from a Neural Network that was trained on CIFAR-10. For ImageNet, K-NN is performed in the 2048 dimensional feature space derived from DenseNet-121.
}

\label{fig:KNN_Analytical_vs_Experimental}
\end{figure}

\begin{figure}[t]
\centering
\begin{subfigure}[MNIST flip]{
\includegraphics[width=0.3\textwidth]{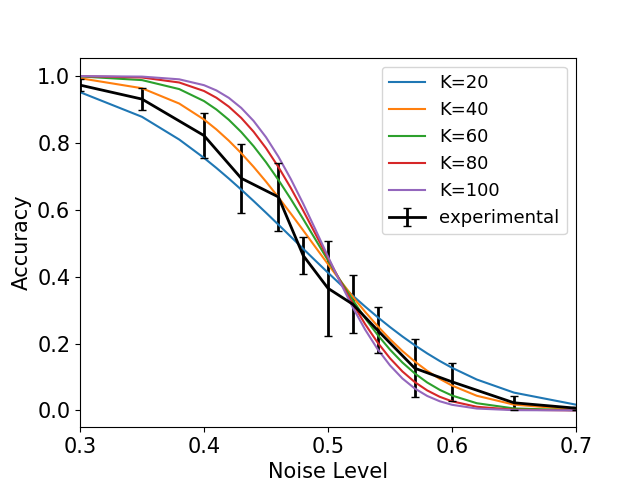}}
\label{fig:MNIST_flip_overlay}
\end{subfigure}
\begin{subfigure}[CIFAR-10 flip]{
\includegraphics[width=0.3\textwidth]{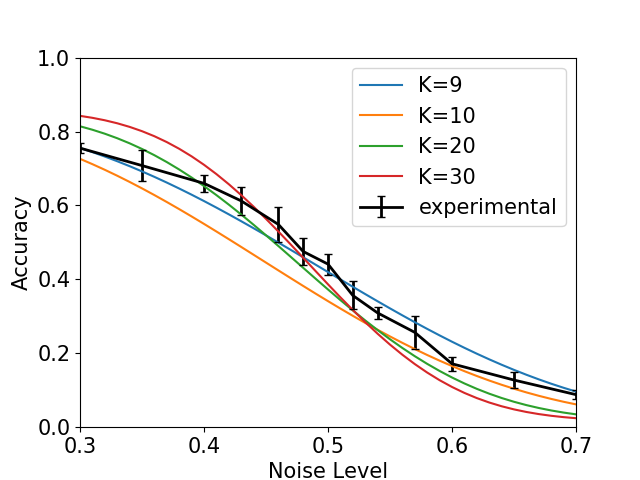}}
\label{fig:CIFAR_flip_overlay}
\end{subfigure}
\begin{subfigure}[ImageNet flip]{ 
\includegraphics[width=0.3\textwidth]{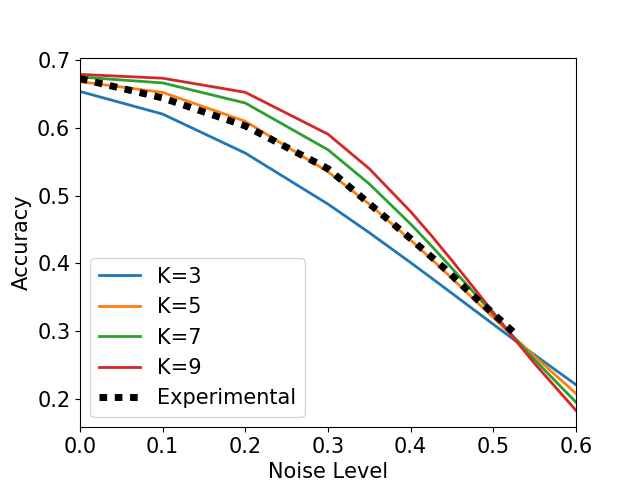}}
\end{subfigure}
\begin{subfigure}[MNIST uniform]{
\includegraphics[width=0.3\textwidth]{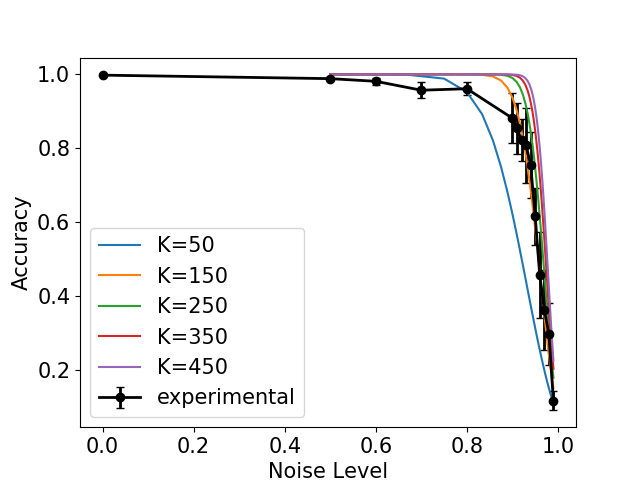}}
\label{fig:MNIST_random_overlay}
\end{subfigure}
\begin{subfigure}[CIFAR-10 uniform]{
\includegraphics[width=0.3\textwidth]{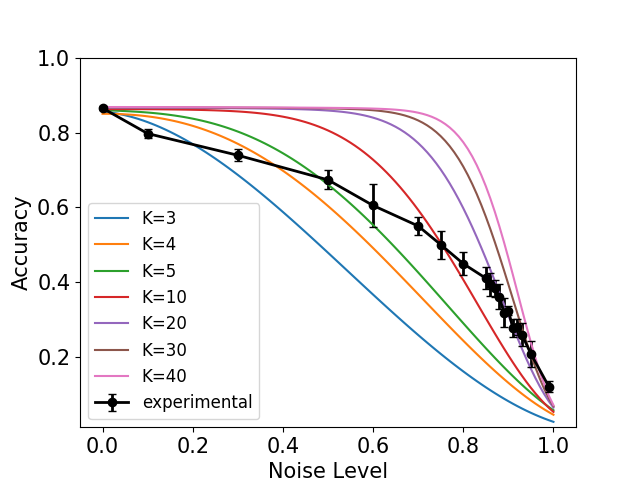}}
\end{subfigure}
\begin{subfigure}[ImageNet uniform]{
\includegraphics[width=0.3\textwidth]{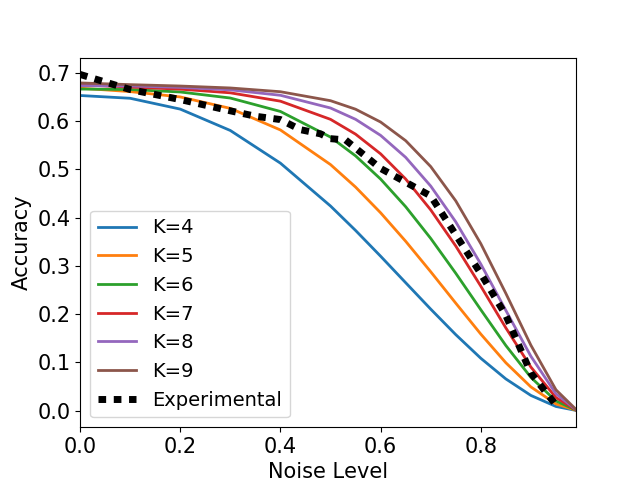}}
\label{fig:ImageNet_random_overlay}
\end{subfigure}
\begin{subfigure}[MNIST general corruption matrix]{
\def\big{\includegraphics[width=0.3\textwidth]{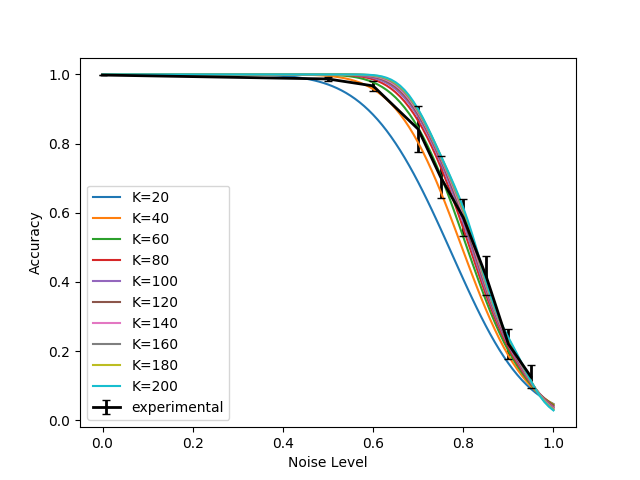}}
\def\little{\includegraphics[width=0.095\textwidth]{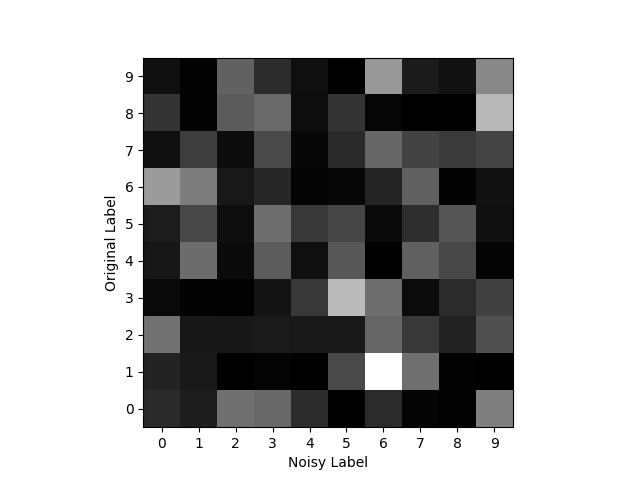}}
\def\stackalignment{l}
\topinset{\little}{\big}{0.1\textwidth}{0.108\textwidth}}
\label{fig:general_corruption_matrix}
\end{subfigure}
\begin{subfigure}[MNIST concentrated noise]{
  \includegraphics[width=0.3\textwidth]{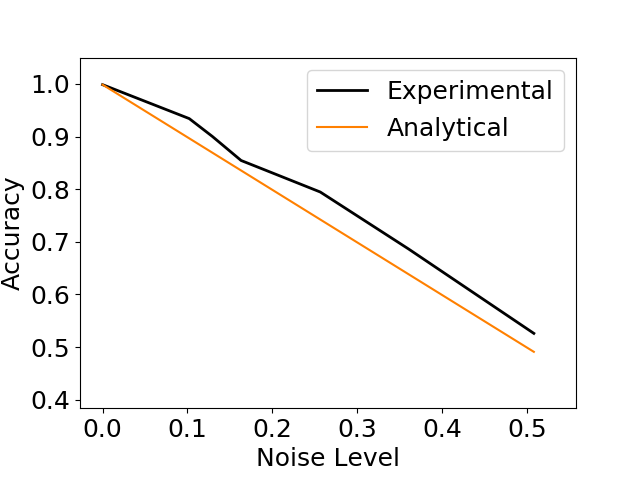}}
\label{fig:mnist_concentrated_curve}
\end{subfigure}%
\begin{subfigure}[CIFAR-10  concentrated noise]{
  \includegraphics[width=0.3\textwidth]{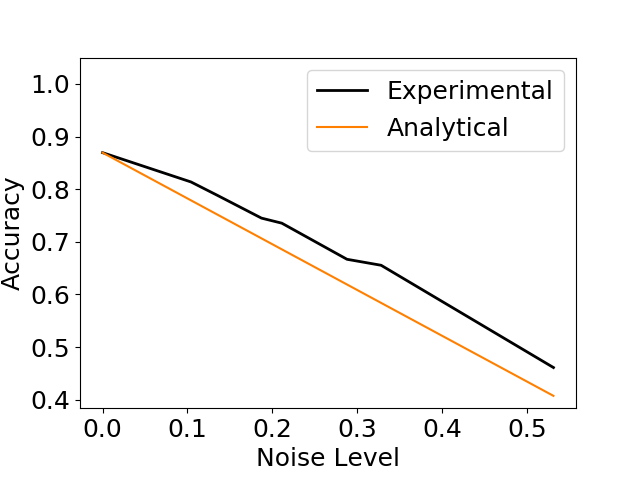}}
\label{fig:cifar_concentrated_curve}
\end{subfigure}
\caption[]
{\small DNN Analytical and Experimental curves. The experimental curves show the mean accuracy and standard deviation. In most cases, the experimental curve is quite close to the corresponding analytical curves, and is clearly different from the analytical curves of the other settings (other subfigures).
In (g) we also show the corruption-matrix  $P(\tilde{y}|y)$ (rows are original label, columns are corrupt label, and brightness denotes probability, where white=high, black=low).  
}
\label{fig:noise_results}
\end{figure}

\noindent {\bf Estimating the clean distribution:} In the $K$-NN setting, we can find the clean distribution by simply analyzing the clean data and noting the labels of the examples in the $K$-neighborhood of each test example. In the DNN setting, we could do the same by selecting a specific embedding space in which to measure distances, e.g. the network's penultimate layer's output. However, such an explicit selection is unnecessary: instead, we follow our observations in Fig.~\ref{fig:softmax} and treat the softmax layer output (of a network trained on clean data) as the clean distribution. This results in a much more computationally efficient algorithm.  

\subsection{The Locally-Concentrated Noise Setting.}
An approximate analysis of DNN accuracy based on the $K$-NN algorithm can be done also in the locally concentrated noise setting. Following our observation that DNN behaves similarly to $K$-NN performed in some implicit space, we need to assume that the noisy examples are concentrated in this space. 

If the noise is concentrated, then $\Nbr(\hat{x})$ is almost always contained either in the \emph{corrupt} area or \emph{clean} area. In the first case, the prediction will be based on the corrupt label, therefore wrong.  In the second, it will be correct. Therefore, the expected accuracy can be determined by the fraction of test examples for which $\Nbr(\hat{x})$ is in the clean area. If we assume that the \emph{test} examples are approximately uniformly spread in the example space, we can expect this fraction to be $1\mkern-4mu-\mkern-2mu\gamma$. Figs.~\ref{fig:softmax}(c,d) and Fig.~\ref{fig:noise_results}(h,i) demonstrate that this is indeed the case empirically.

\section{Experiments} \label{sec:experiments}

Our analytical model for $K$-NN acurracy in the presence of noise provides accuracy-vs-noise curves. We compare these to experimental curves derived from performing $K$-NN and DNN on noisy data. We repeat these experiments with multiple noise types, and several popular datasets: MNIST, CIFAR-10, and ImageNet (ILSVRC 2012). Notice that we re-train the network for each dataset, noise type and noise level. 

{\bf $K$-NN experiments:}
Fig.~\ref{fig:KNN_Analytical_vs_Experimental} shows how well Eq.~\eqref{eq:Q} approximates K-NN classification in practice. We show results for both MNIST and CIFAR-10 for the case of uniform and flip label noise, and for different $k$ values. As can be seen, our model fits the data well. \\
For each dataset we use an embedding of the examples in a space that is semantically meaningful enough to allow successful classification using $K$-NN. The MNIST dataset is simple enough that we can simply use the image pixels.  For the other datasets, however, we use the output of the penultimate layer of the DNN trained to classify the dataset (described hereafter). 

{\bf Implementation details:}
The analytical expressions in Eq.~\eqref{eq:efficient_nested_summation} are computationally intensive. For a feasible run-time, we found it necessary to use a multi-threaded C++ implementation that relies heavily on dynamic programming. On a fast 8-core Intel i7 CPU, producing the analytical curves for each figure takes up-to 60 minutes. 

To generate the empirical plots, we repeat the following process for different noise levels: add noise to the training set, train a DNN, and then measure its accuracy on the clean test set. When it is computationally feasible, we repeat the experiment several times and estimate the mean accuracy and its standard deviation (10 repeats for MNIST, 7 for CIFAR-10). 
For CIFAR-10 and MNIST we use a train/validation split of 90\%/10\%. The clean validation set is used for early stopping~\cite{plaut86}. 
We perform this in our experiments to improve the network performance following the observations in  \cite{Arpit17Closer, Carlini19Secret}: one effect of overfitting may be memorization of noisy labels, which could degrade DNN's resistance to label-noise.

For all MNIST experiments, we use a DNN, which reaches  $\sim$100\% accuracy. Its structure is described in the sup. material. 
For the CIFAR-10 experiments, we use the All Convolutional Network \cite{DBLP:journals/corr/SpringenbergDBR14}. To produce features for the $K$-NN experiments, an additional fully connected layer was added before the softmax, with 256 output channels.
 For ImageNet experiments, we use the Densenet-121 \cite{DBLP:journals/corr/HuangLW16a} architecture, with Adam Optimization and mini-batch of size $256$. The feature used in $K$-NN experiments is 2048-dimensional.

The results of our experiments are summarized in Fig.~\ref{fig:noise_results}. They contain four types of noise (uniform, flipped, general confusion matrix, and locally concentrated), three datasets (MNIST, CIFAR-10, and ImageNet), and different values of $K$ (\ie, different neighborhood sizes). To experiment with locally concentrated noise, we need to explicitly select a specific space and select groups of examples that are concentrated in this space. Specifically, we use the output of the penultimate layer of a network trained on clean data as a $256$-dimensional embedding space. We use $k$-means to find clusters of examples that are locally-concentrated in this space.  For each class separately: we divide into $k$ clusters, then we select one cluster  and change all of the labels in it into the same incorrect label. Each class $\ell_i$ has one alternative class $\ell_j$ to which the noisy labels are flipped. $k$-means with different values of $k$ results in different noise-levels, from roughly 10\% when $k=10$, to roughly 50\% when $k=2$.

{\bf Validation of analytical expression:} The graphs show that we can calculate analytically the performance of the network for a given noise level, for some types of label noise. Specifically, the black line in each graph shows the performance of a network trained  with a growing amount of label noise. The colored line curves show graphs computed analytically that determine the performance of a network, given different neighborhood sizes. That is, we show that there is a connection between label noise and neighborhood size. This connection lets us compute analytically the expected accuracy of a network without having to train it.
In all cases, the experimental curve appears to naturally follow its corresponding family of analytical curves. We believe this indicates that the analytical curves approximate the \emph{general behavior} of the experimental curves. In other words, our mathematical analysis captures a \emph{major} factor in explaining the resistance of DNNs to spatially-spread noise. On a smaller scale, there are some deviations of the experimental curves from the anlytical ones. This could be caused by secondary factors that are not considered by the model. 


{\bf The impact of noise concentration on DNN's resistance to label noise:} The analytical expression predicts that DNNs may resist high levels of noise, but only if the noise is \emph{randomly spread} in the training set (i.e., the uniform and flip settings). In contrast, in the locally concentrated noise setting DNNs are expected to have no resistance to noise. Note that indeed, this predicted behavior is demonstrated in the plots. In particular, our experiments show that the \emph{uniform} noise setting is easier for the network than the \emph{flip} setting. In the flip case, resistance to noise holds only until the noise level approaches 50\%. In the uniform noise setting, noticeable drop in accuracy happens only when approaching 90\%. This is due to the fact that in the flip setting, at 50\% there is a \emph{reversal of roles} between the correct label and the alternative labels, and the DNN ends up learning the alternative labels while ignoring the correct ones. In the uniform noise setting, however, the probability of the correct label being the plurality label is still higher than that of any of the other labels. Note that all these behaviours are in accordance with our theory.

\section{Conclusion}

This work studies the robustness of $K$-NN and neural networks to label noise. We provide an analytic expression that predicts accurately the resistance of $K$-NN to label noise and can be evaluated efficiently. Moreover, we show that the developed formula can also be used to understand the behavior of neural networks in the presence of different noise types.  

The underlying assumption for using this term also for trained neural networks is that they behave similarly to $K$-NN. This assumption is related to recent studies of the function space of neural networks that show that network perform a spline interpolation between training examples \cite{Ongie2020Function,Savarese19How,Williams19Gradient,giryes2020function}. $K$-NN can be viewed as a first-order approximation of such an interpolation, especially in the classification case, which is studied in this work. Yet, the value of $K$ that provides this approximation changes between experiments. An open question is how to set the value of $K$. Future work may pursue this direction, or suggest how to incorporate additional factors into the approximation of error bounds for DNN with label noise. 

The relation between DNN and $K$-NN is especially evident in their performance when trained with noisy data. 
We performed several experiments that demonstrated this intuition and then compared empirical results of training neural nets with label noise, with analytical (or numeric) curves derived from a mathematical analysis of the $K$-NN model. Indeed, the analytic curves are less accurate in the DNN case, as they provide a first order approximation. Yet, in many cases they capture the expected behavior of DNNs. In particular, they show that DNN robustness to label noise depends on concentration of noise in the training set. This explains the incredible resistance of DNNs to spatially spread noise and their degradation in performance in the case of locally concentrated noise. 

Notice that our findings can be used to provide new insights on the techniques that are used to improve neural network robustness to label noise. For example, one popular approach is to incorporate the confusion matrix into the training
\cite{Mnih12Learning,DBLP:conf/cvpr/XiaoXYHW15,Sukhbaatar15Training, Goldberger17Training, Patrini17Making}. Taking a distillation perspective, training with a confusion matrix can be viewed as training a student network to imitate a teacher network that  was trained regularly just using the plain labels. This follows from our observation in this work (see for example Fig.~\ref{fig:softmax}) that the average histogram at the output of the network when trained with the noisy labels resemble the confusion matrix of the data labels. To the best of our knowledge, we are the first to show this phenomenon.  Future work may use this observation as well as other observations presented in this work to develop new methods for robustness to label noise.

\section*{Acknowledgments} 
This work is partially supported by the by the IIA MDM Consortium. This work was partly funded by ISF grant number 1549/19. This research was supported by Trax Image Recognition for Retail and Consumer Goods.

\bibliography{DnnNoise}

\appendix
\pagebreak
\title{Appendix}

\appendix

\section{Contents of Appendix}
The appendix contains the following:
\begin{itemize}
    \item An experiment demonstrating the similarity between the softmax output of a network, and $K$-NN performed in the space of the outputs of its one-before-last layer. 
    \item Derivation of algorithm for efficient calculation of the analytical expression for the expected accuracy of $K$-NN in the presence of label noise.
    \item Simplified analysis of the anlytical expression for special cases.
    \item Additional implementation details.
    \item Diagrams demonstrating softmax. outputs for networks trained on various datasets with different noise models.
\end{itemize}

\section{Comparison of Softmax Outputs to K-NN Histograms}\label{sec:SoftmaxVsKnn}
\begin{figure*}
\centering
\begin{subfigure}{
\includegraphics[width=0.475\linewidth]{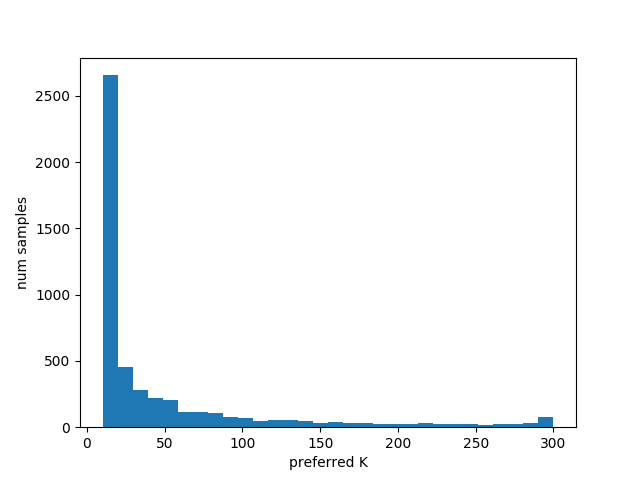}}
\end{subfigure}
\hfill
\begin{subfigure}{
\includegraphics[width=0.475\linewidth]{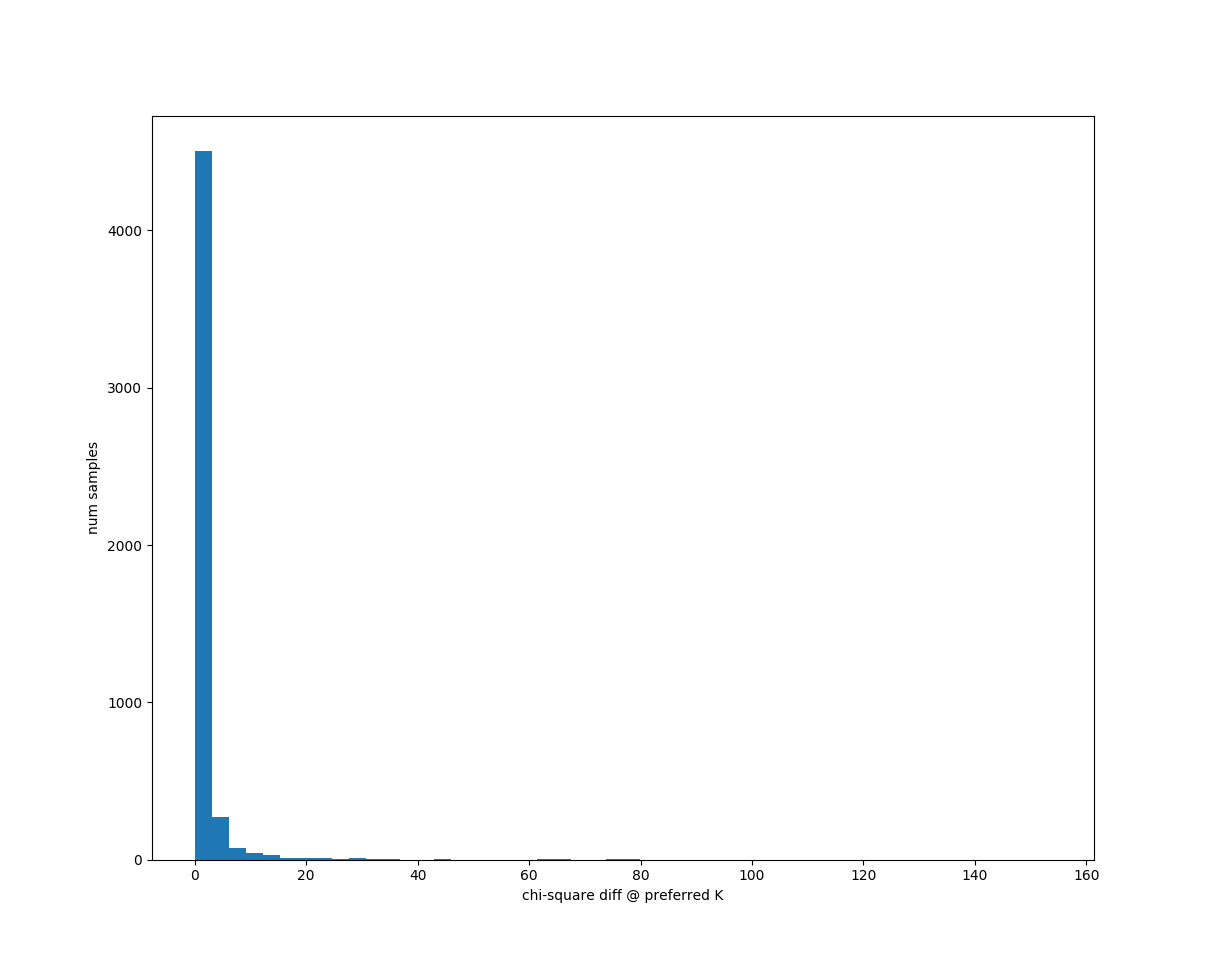}}  
\end{subfigure}
\caption{For each test example its \emph{preferred $K$} is the value of $K$ which yields the lowest chi-square distance to the softmax layer output. (a) shows a histogram of the prevalence of different choices of $K$. (b) shows a histogram of the chi-square distances, when each example is at its preferred $K$ }    
\label{fig:chiSquareHistograms}
\end{figure*}

In this work, we have presented the conjecture that the output of the softmax layer tends to encapsulate the local distribution of the train examples in the vicinity of a given test example. In the main paper we demonstrate this by injecting noise into the training set, without having to explicitly define the space in which $K$-NN operates. Here, we demonstrate the similarity for a specific space: the 256-dimensional output of the penultimate layer of a network trained on clean data. We produce histograms of labels for K-Nearest Neighbors (with different values of K), and calculate the chi-square distance from these histograms to the softmax layer output. The network used for the embedding space is trained on a clean version of the CIFAR10 dataset, and has the following structure: cnv@20 - cnv@20 - pool - cnv@50 - cnv@50 - pool - fc@\emph{256} - fc@10 - softmax, \\
where \emph{cnv} is a convolutional layer using a 5x5 filter and zero-padding, \emph{fc} is a fully connected layer, \emph{@c} denotes the number of output channels, and \emph{pool} is 2x2 max-pooling . Batch Normalization is added after each convolutional and fully-connected layer, followed by a ReLU non-linearity (except before the softmax layer). The features we use are the raw outputs of the fully connected layer with 256 output channels, before they are passed into batch normalization and ReLU. We try a range of K values, between 10 and 300, and for each example select its \emph{preferred K value}, which is the one with the lowest chi-square distance. Fig.~\ref{fig:chiSquareHistograms}(a) shows the prevalence of different choices of K. Fig.~\ref{fig:chiSquareHistograms}(b) presents the histogram of the calculated chi-square distances. 

\begin{figure*}
\centering
\begin{subfigure}{
\includegraphics[width=0.475\textwidth]{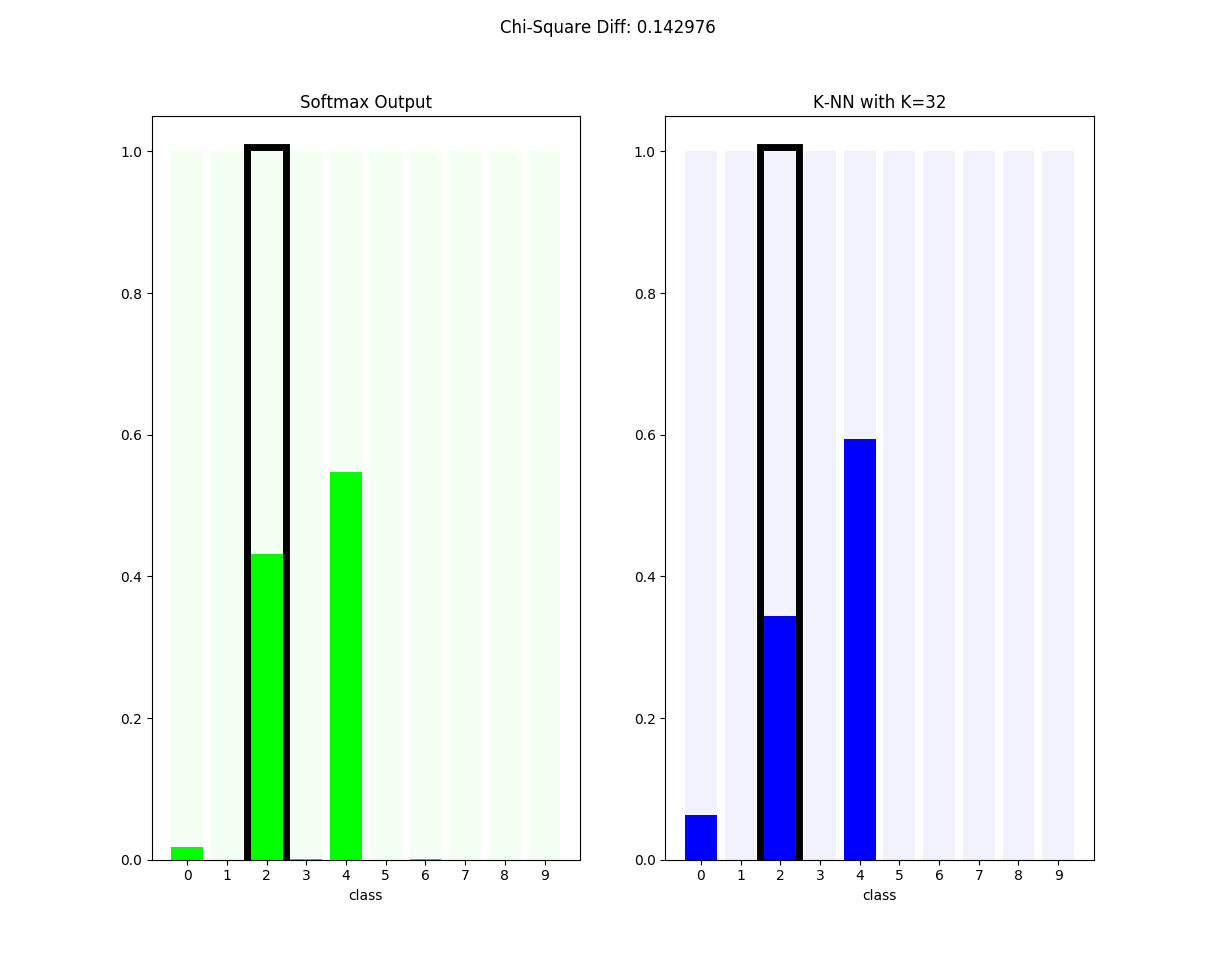}}
\end{subfigure}
\hfill
\begin{subfigure}{
\includegraphics[width=0.475\textwidth]{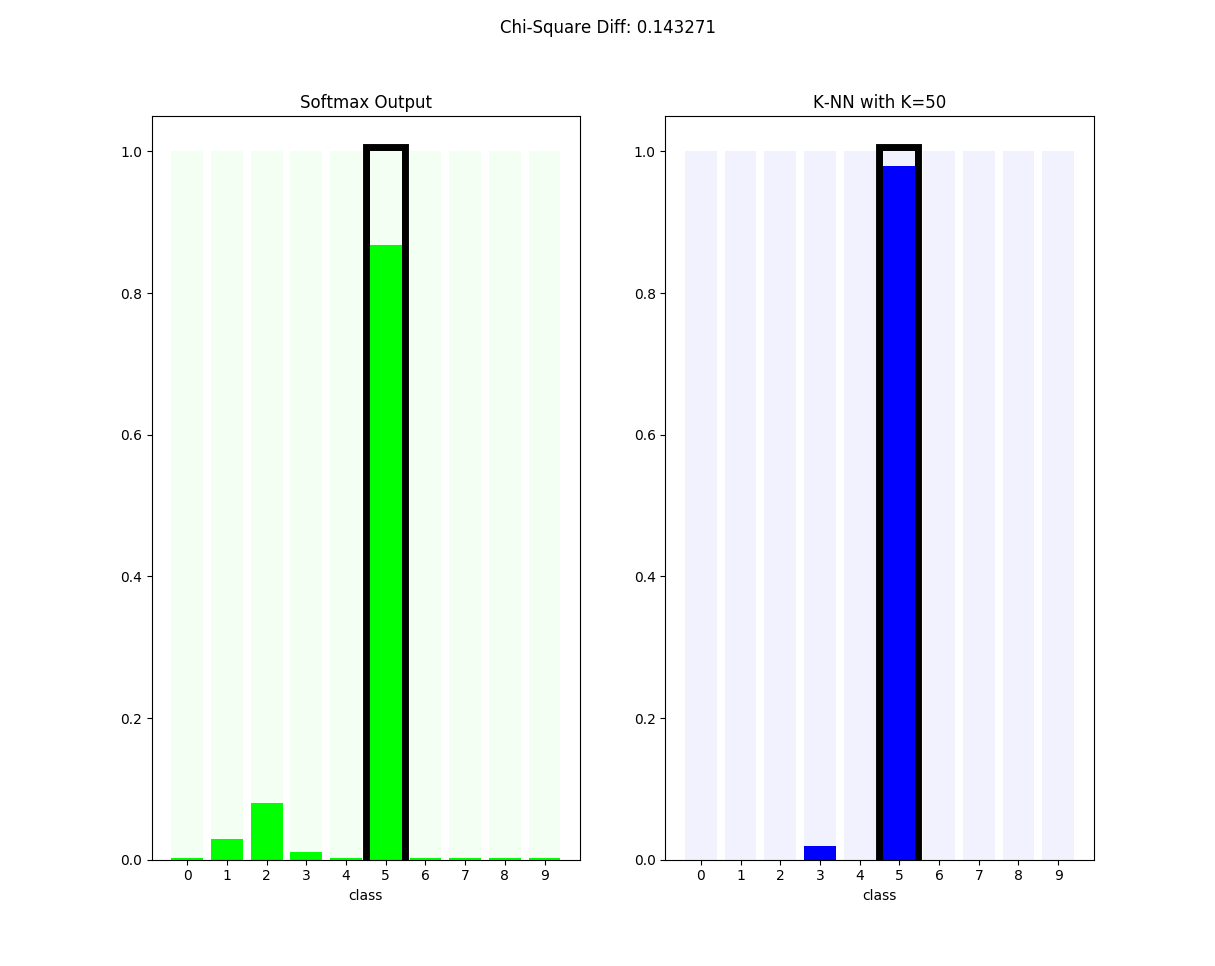}}
\label{fig:RandomNoise}
\end{subfigure}
\vskip\baselineskip
\begin{subfigure}{
\includegraphics[width=0.475\textwidth]{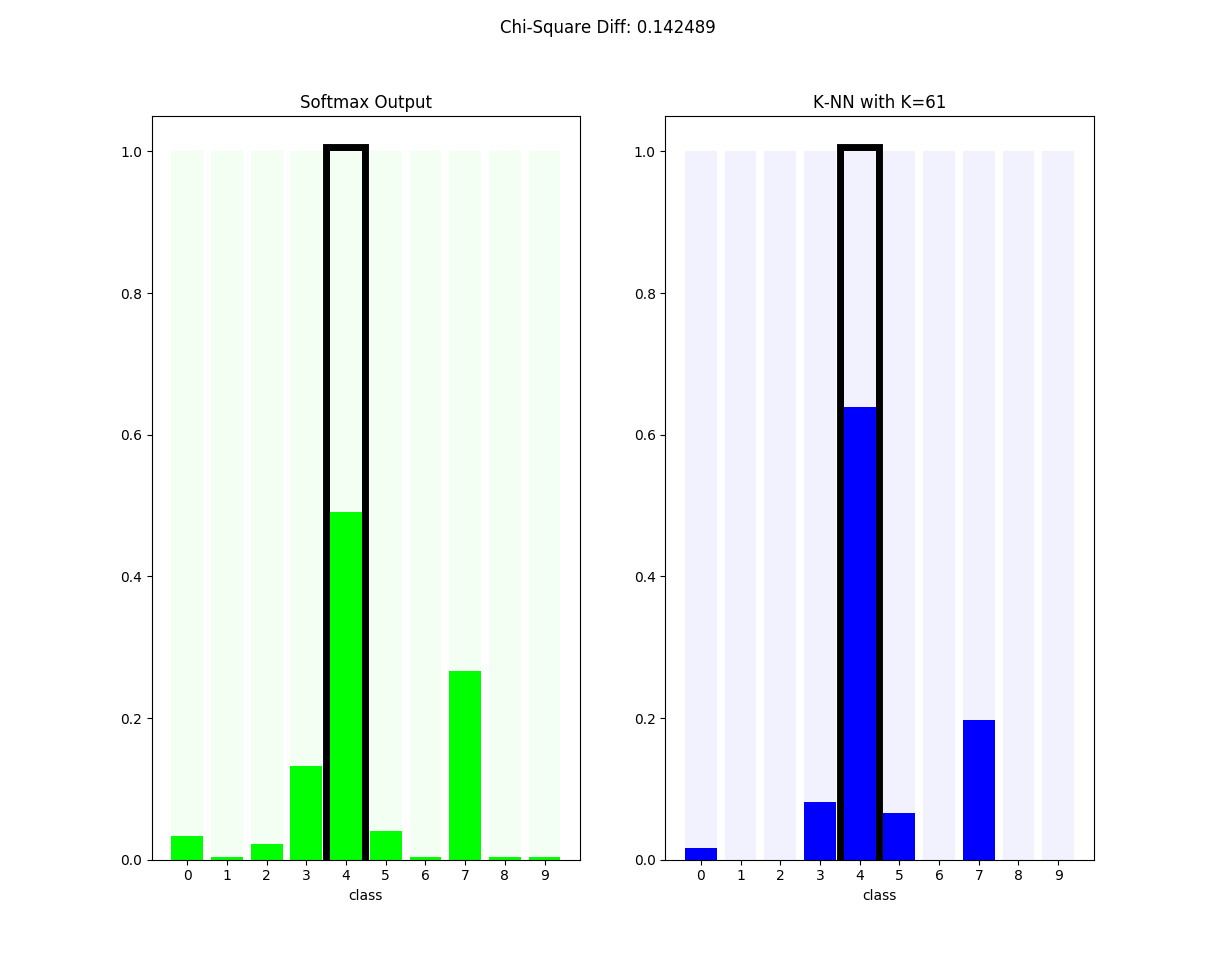}}
\end{subfigure}
\hfill
\begin{subfigure}{
\includegraphics[width=0.475\textwidth]{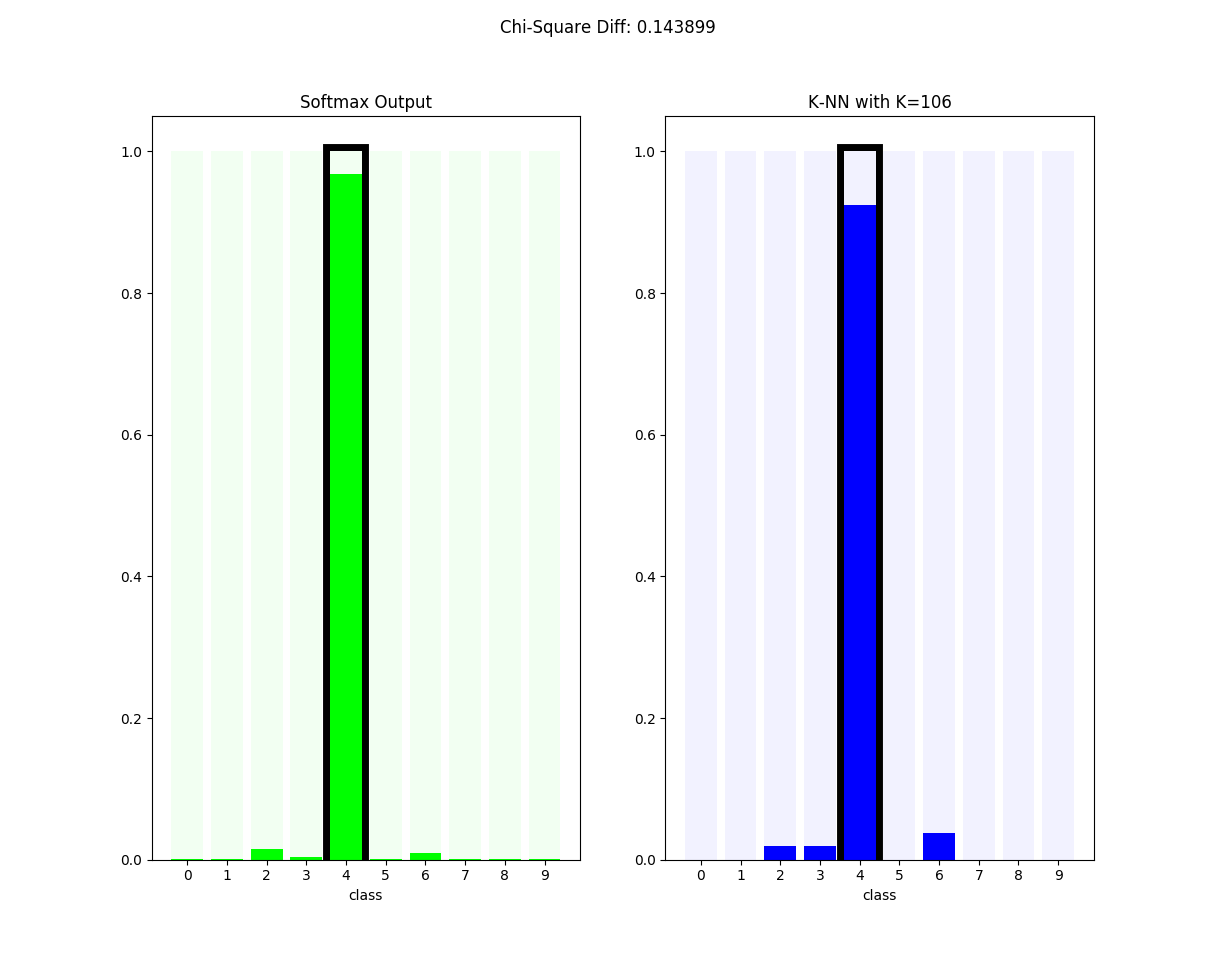}}
\label{fig:mean and std of net44}
\end{subfigure}
\caption{The median chi-square distance between softmax layer output and K-NN histogram is 0.143123. To get a sense of the meaning of this number, we show a comparison of histograms for several examples where the chi-square distance is around this value. In each pair the left (green) histogram is the softmax layer output, and the right (blue) is the K-NN histogram for the example's preferred K.}
\label{fig:chi_median}
\end{figure*}

The median chi-square distance between softmax layer output and K-NN histogram is $0.143123$, which shows that the distributions are very close to each other. 
To get a better sense of the meaning of this number, we show a comparison of histograms for several examples in Fig.~\ref{fig:chi_median}, where the chi-square distance is around this value. In each pair, the softmax output and the K-NN histogram for the example's preferred K are presented. It can bee seen that these histograms are very close to each other. 

\section{Efficient Calculation of The Analytical Expression \label{sec:efficent_summation}}

We turn to present here an efficient strategy for computing the probability $Q$ in Theorem 1 in the main paper. 
A naive computation of it, may iterate over all possible combinations of $n_1, n_2, \ldots$, but only sum those where the plurality label is the correct one. As we shall see now, in addition to being inefficient, this is also unnecessary. 

To make the calculation more efficient, we calculate the lower and upper boundaries of each $n_i$ such that the summation only goes through the combinations that lead to a correct plurality label.  Denoting the lower bounds by $m_i$ and the upper bounds by $M_i$, we have that
\begin{align}
\label{eq:Q_ni_mi_Mi1}
Q &= \sum_{n_1=m_1}^{M_1}\sum_{n_2=m_2(n_1)}^{M_2(n_1)}\cdots\sum_{n_L=m_L(n_1,\ldots,n_{L-1})}^{M_L(n_1,\ldots,n_{L-1})} \nonumber \\
& \binom{K}{n_1,n_2,\ldots,n_L}q_1^{n_1}\cdot q_2^{n_2}\cdot\,\cdots\,\cdot q_L^{n_L},
\end{align}
where $m_i$ is the smallest number of repeats of $\ell_i$ allowed, and $M_i$ is the largest one. Their possible values are calculated in Section~\ref{sec:mi_Mi_def}. Notice that the number of repeats allowed for any label $\ell_i$ depends on the number of repeats already selected for all the previous labels, $\ell_j \,\,\forall j<i$. 

For further efficiency, we can now decompose the summed expression so that shared parts of the calculation are only performed once. We decompose the multinomial coefficient into a product of binomial coefficients as follows:
\begin{align}
\binom{K}{n_1,n_2,\ldots,n_L} & =   \\ \nonumber
& \hspace{-0.2in} \binom{K}{n_1}\cdot\binom{K-n_1}{n_2}\cdots\binom{K-\sum_{j=1}^{L-1}n_j}
{n_L},
\end{align}
and get the following formula for calculating $Q$:
\begin{align}
 Q &= \sum_{n_1=m_1}^{M_1}{\binom{K}{n_1}}q_1^{n_1}
\cdots \mkern-9mu \sum_{n_L=m_L}^{M_L}{\binom{K-\sum\limits_{j=1}^{L-1}n_j}{n_L}}q_L^{n_L}.
\end{align}


\subsection{Defining $m_i$ and $M_i$}
\label{sec:mi_Mi_def}
We will assume, without loss of generality, that the correct label is $\ell_1$. Clearly, we can repeat the same analysis by  simply renaming or shuffling the labels. $m_i$ and $M_i$ need to be defined in a way that ensures:
\begin{enumerate}
\item There are exactly K letters in the string.
\item $\ell_1$ is the plurality label, i.e. $n_1 > n_i \,\,\,\forall i \neq 1$.
\end{enumerate}
We can start with $M_1$, which is simply $K$. Clearly, a string consisting of K repeats of $\ell_1$ fulfills both requirements. Once $n_1$ is known, we can define the maximum allowed number of repeats for any other letter as $M^* = n_1 -1$. 
With the definition of $M^*$, we turn to calculate $m_1$. 
Since $\sum_i n_i = K$ and $n_i\le M^*$, we have that 
\begin{eqnarray}
K \le n_1 + (L-1)M^* =  n_1 + (L-1)(n_1 - 1).
\end{eqnarray}
By reordering the terms, we get that
\begin{align}
\label{eq:n_1_K_inequality}
n_1  \geq  \frac{K + (L-1)}{L}. 
\end{align}
Using the fact that $m_1$ is the smallest integer satisfying \eqref{eq:n_1_K_inequality}, we have
\begin{align}
 m_1 = \left\lceil\frac{K + (L-1)}{L}\right\rceil.
\end{align}


Having $m_1$ and $M_1$ set, we turn to calculate the values of  $M_i \,\,\forall i \neq 1$. We start by defining $R_i$ which is the number of string positions that are still unassigned:
\begin{equation}
R_i = K - \sum_{j=1}^{i-1}n_j.
\end{equation}
Clearly, the value of $n_i$ should be no larger than $R_i$. Thus,
\begin{equation}
M_i = \min\{R_i, M^*\}.
\end{equation}
Lastly, we define $m_i$ in a way that makes sure the string has no less than K letters:
\begin{equation}
m_i = \max\{0, R_i-(L-i)\cdot M^*\}.
\end{equation}
The intuition here is that if all the subsequent letters $\ell_{i+1},\ldots,\ell_L$ have the maximal number of repeats, $M^*$, then $\ell_i$ need to be repeated enough times to bring the total repeats of all the yet unassigned letters to $R_i$.  

\section{Simplified analysis of special cases} \label{sec:simplified}

The process of calculating $Q$ can be accelerated by several orders of magnitude if the following requirements are met:
\begin{enumerate}
\item The dataset is almost \emph{perfectly learnable}, meaning that a CNN is able to reach approximately 100\% test accuracy when trained with clean labels. 
\item The conditional probabilities $P(\tilde{y}|y)$ are the same for all $y$, up to renaming of the labels. 
\item The distribution of labels in the test set is \emph{balanced}, meaning there is the same number of test examples for each label.
\end{enumerate}

In these cases, the perfect learnability allows us to simplify $C$ by assuming that for all train examples $x$, \emph{all} clean labels in $\Nbr(\hat{x})$ are the correct label:
\begin{equation}
C(\ell) = \begin{cases}
1 & \ell=\hat{y}\\
0 &\text{else}
\end{cases}
\end{equation}
Also, the probability $Q$ is the same for all test examples, from which follows $A_{K-NN}=Q$. 
For the \emph{uniform noise} setting, $q_j$ is simplified to
\begin{equation}
q_j  = \begin{cases}
(1\mkern-4mu-\mkern-2mu\gamma) + \frac{\gamma}{L} & \ell_j=\hat{y}\\
\frac{\gamma}{L} &\text{else},
\end{cases}
\end{equation}
and for the \emph{flip noise} setting, $Q$ is simplified to
\begin{equation}
Q = \Pr \big( Y(\hat{x}) = \hat{y} \big) = \sum_{n=\ceil{\frac{K+1}{2}}}^{K} \binom{K}{n}\cdot (1\mkern-4mu-\mkern-2mu\gamma)^{n}\cdot \gamma^{K-n}, \label{eq:Q_flip}
\end{equation}
where $n$ is the number of examples in $\Nbr(\hat{x})$ that have not been corrupted, and $K-n$ is the number of those that have been corrupted, i.e. flipped to the alternative label.

\section{Additional Implementation Details} \label{sec:additional_implementation_details}

For all MNIST experiments, we use a DNN inspired by LENET-5 \cite{Lecun98gradient-basedlearning} and AlexNet \cite{NIPS2012_4824}, which reaches  $\sim$100\% accuracy. Its structure is: cnv@20 - cnv@20 - pool - cnv@50 - cnv@50 - pool - fc@\emph{FS} - fc@10 - softmax, where \emph{cnv} is a convolutional layer using a $5\times5$ filter and zero-padding, \emph{fc} is a fully connected layer, \emph{@c} denotes the number of output channels, and \emph{pool} is $2\times2$ max-pooling. \emph{FS} is 500 for Uniform Noise experiments and 256 for Flip Noise experiments. Batch Normalization \cite{Ioffe:2015:BNA:3045118.3045167} is added after each convolutional and fully-connected layer, followed by a ReLU non-linearity (except before the softmax layer). 

Our data pre-processing in ImageNet training is inspired by ResNet \cite{He2016DeepRL}. Each image is resized so that its shorter side is changed to $256$ (and the rest maintain the same aspect ratio). 
For training, we randomly sample a $224\times 224$ crop from an image. For the test set we simply take the crop from the center of each image. As the network architecture, we use Densenet-121 \cite{DBLP:journals/corr/HuangLW16a} with Adam Optimization and mini-batch of size $256$. The learning rate is
initiated to $0.001$ and then divided by $10$ after $15$ epochs.
The models are trained up-to $30$ epochs with early stopping. \\
For running-time considerations, The $K$-NN experiments on ImageNet were done using not the entire train set but instead a randomly selected subset of 2000 test examples (out of 50000 total).

\begin{figure}
\centering
\begin{subfigure}[Clean]{
  \includegraphics[width=0.21\linewidth]{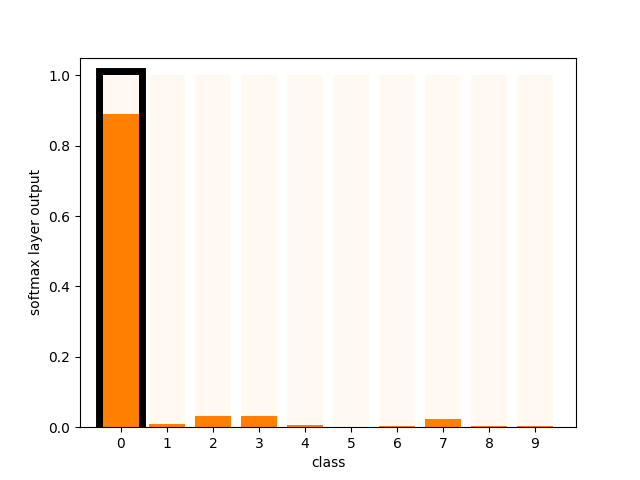}}
\label{fig:cifar_bar_concentrated_clean}
\end{subfigure}%
\begin{subfigure}[30\% uniform noise]{
  \includegraphics[width=0.21\linewidth]{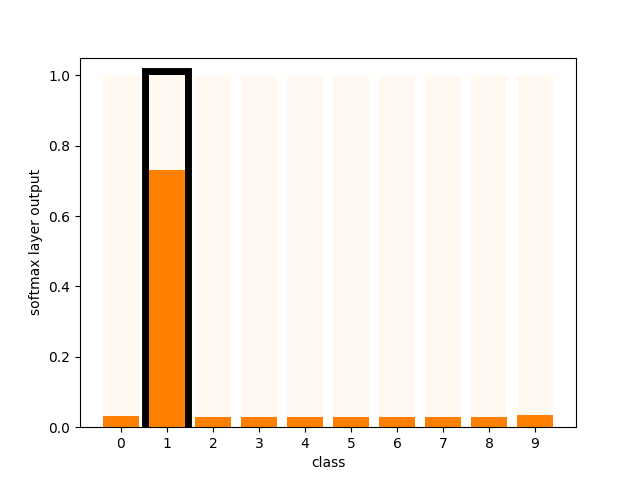}}
\label{fig:cifar_bar_uniform_30}
\end{subfigure}
\begin{subfigure}[60\% uniform noise]{
  \includegraphics[width=0.21\linewidth]{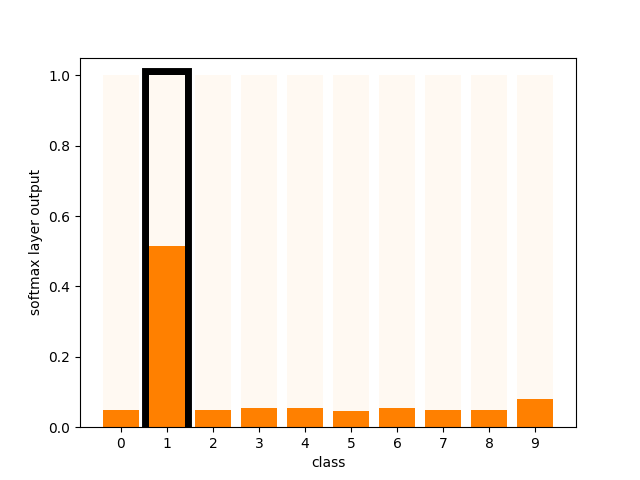}}
\label{fig:cifar_bar_uniform_60}
\end{subfigure}
\begin{subfigure}[20\% flip noise]{
  \includegraphics[width=0.21\linewidth]{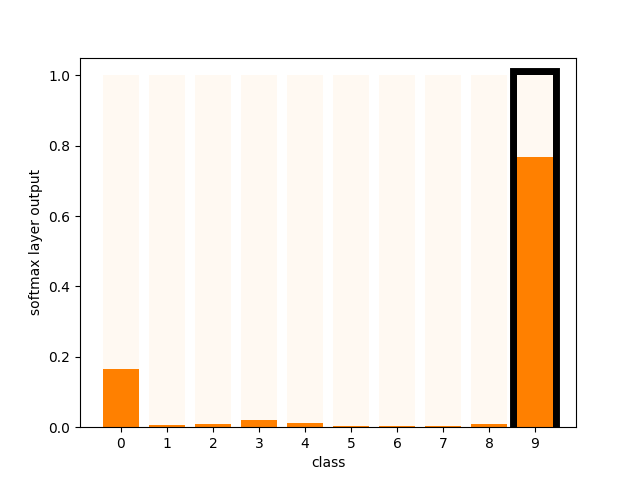}}
\label{fig:cifar_bar_flip_20}
\end{subfigure}
\begin{subfigure}[40\% flip noise]{
  \includegraphics[width=0.21\linewidth]{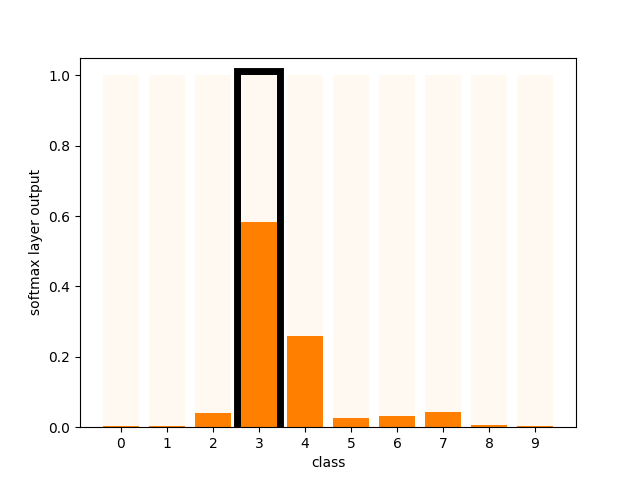}}
\label{fig:bar_flip_40}
\end{subfigure}
\begin{subfigure}[Corruption matrix based noise, with two alternative labels]{
  \includegraphics[width=0.21\linewidth]{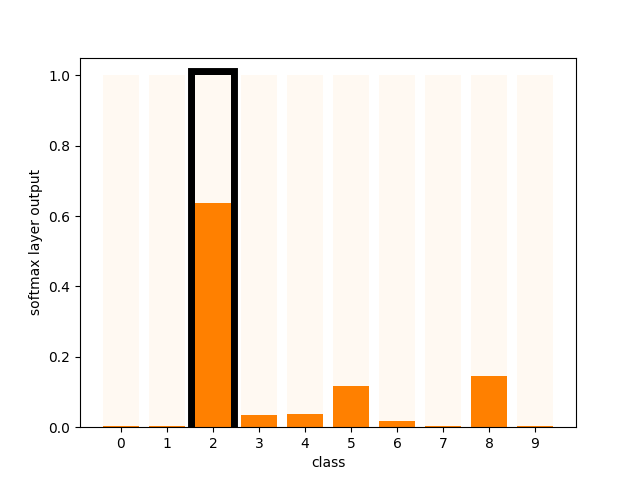}}    
\label{fig:cifar_bar_confusion}
\end{subfigure}
\begin{subfigure}[Locally Concentrated Noise, example in a clean region]{
  \includegraphics[width=0.21\linewidth]{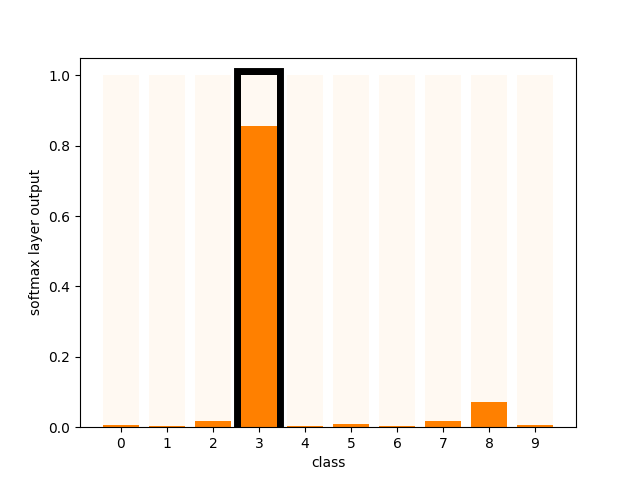}}    
\label{fig:cifar_bar_concentrated_clean2}
\end{subfigure}
\begin{subfigure}[Locally Concentrated Noise, example in a noisy region]{
  \includegraphics[width=0.21\linewidth]{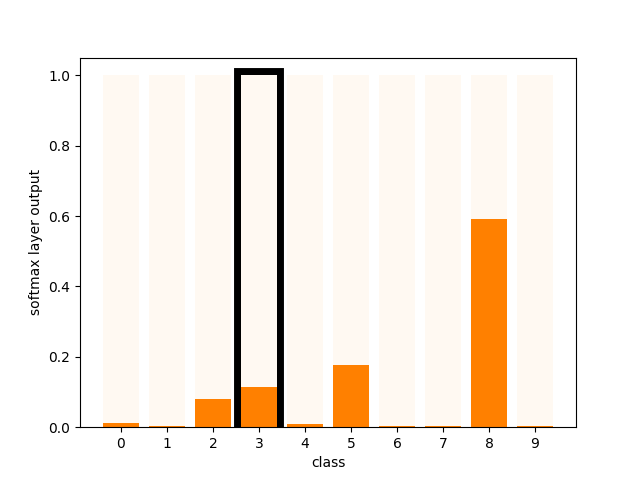}}
\label{fig:cifar_bar_concentrated_noisy}
\end{subfigure}
\caption{\small Softmax outputs of networks trained on noisy versions of the CIFAR-10 dataset. The ground truth label is marked by a black margin. Note that the network output tends to encapsulate the local distribution of labels in the vicinity of the input $x$. }
\label{fig:CifarBarGraphs}
\end{figure}

\begin{figure*}
\centering
\includegraphics[width=\linewidth]{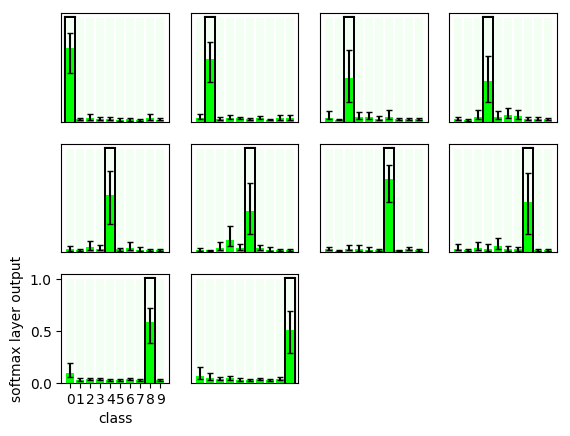}
\caption[]
{\small Detailed Aggregate Softmax outputs for a network trained on CIFAR-10 with 30\% uniform noise. In each diagram, we show an aggregate of softmax vectors taken from all test examples that share the same ground truth label. In the top left diagram the GT label is 0, in the next diagram it is 1, etc. The height of the bars show the median, and the confidence interval shows the central 50\% of examples. The ground truth label is marked by a black margin.}
\label{fig:CIFAR_bar_uniform_30_all}
\end{figure*}

\begin{figure*}
\centering
\includegraphics[width=\linewidth]{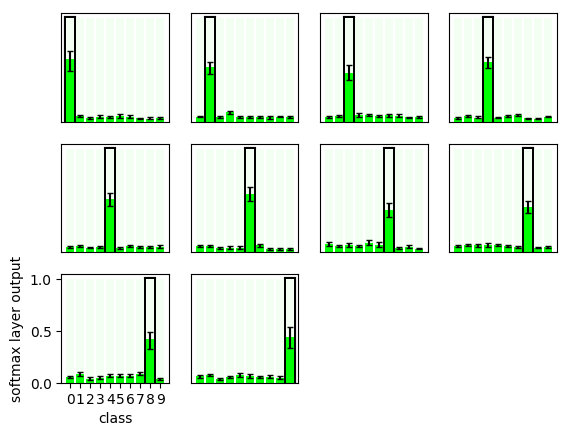}
\caption[]
{\small Detailed Aggregate Softmax outputs for a network trained on MNIST with 60\% uniform noise. In each diagram, we show an aggregate of softmax vectors taken from all test examples that share the same ground truth label. In the top left diagram the GT label is 0, in the next diagram it is 1, etc. The height of the bars show the median, and the confidence interval shows the central 50\% of examples. The ground truth label is marked by a black margin.}
\label{fig:MNIST_bar_uniform_60_all}
\end{figure*}

\begin{figure*}
\centering
\includegraphics[width=\linewidth]{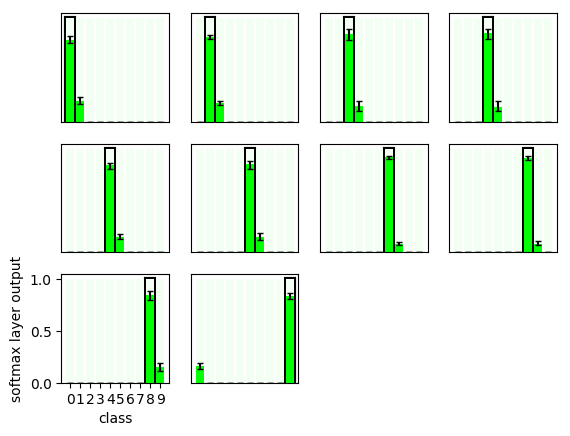}
\caption[]
{\small Detailed Aggregate Softmax outputs for a network trained on MNIST with 20\% flip noise. In each diagram, we show an aggregate of softmax vectors taken from all test examples that share the same ground truth label. In the top left diagram the GT label is 0, in the next diagram it is 1, etc. The height of the bars show the median, and the confidence interval shows the central 50\% of examples. The ground truth label is marked by a black margin.}
\label{fig:MNIST_bar_flip_20_all}
\end{figure*}

\begin{figure*}
\centering
\includegraphics[width=\linewidth]{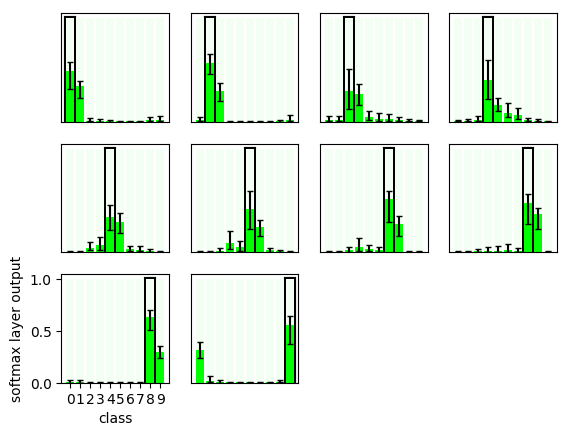}
\caption[]
{\small Detailed Aggregate Softmax outputs for a network trained on CIFAR with 40\% flip noise. In each diagram, we show an aggregate of softmax vectors taken from all test examples that share the same ground truth label. In the top left diagram the GT label is 0, in the next diagram it is 1, etc. The height of the bars show the median, and the confidence interval shows the central 50\% of examples. The ground truth label is marked by a black margin.}
\label{fig:CIFAR_bar_flip_40_all}
\end{figure*}

\begin{figure*}
\centering
\includegraphics[width=\linewidth]{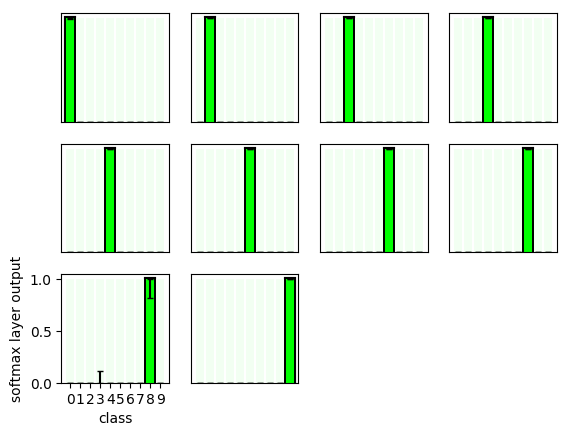}
\caption[]
{\small Detailed Aggregate Softmax outputs for a network trained on MNIST with 25\% locally concentrated noise, with an example in the clean region. In each diagram, we show an aggregate of softmax vectors taken from all test examples that share the same ground truth label. In the top left diagram the GT label is 0, in the next diagram it is 1, etc. The height of the bars show the median, and the confidence interval shows the central 50\% of examples. The ground truth label is marked by a black margin.}
\label{fig:MNIST_bar_concentrated_clean_all}
\end{figure*}

\begin{figure*}
\centering
\includegraphics[width=\linewidth]{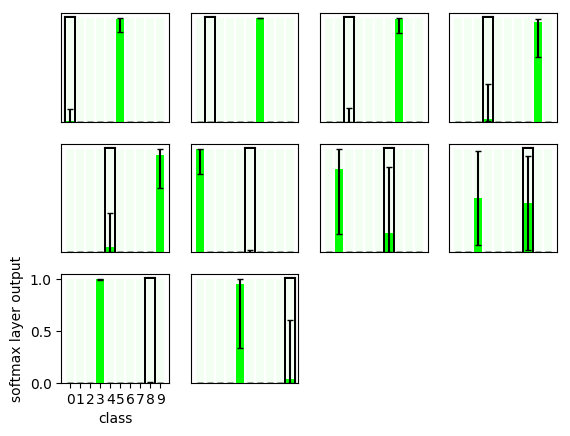}
\caption[]
{\small Detailed Aggregate Softmax outputs for a network trained on MNIST with 25\% locally concentrated noise, with an example in the \emph{noisy} region. In each diagram, we show an aggregate of softmax vectors taken from all test examples that share the same ground truth label. In the top left diagram the GT label is 0, in the next diagram it is 1, etc. The height of the bars show the median, and the confidence interval shows the central 50\% of examples. The ground truth label is marked by a black margin.}
\label{fig:MNIST_bar_concentrated_noisy_all}
\end{figure*}


\end{document}